\documentclass[conference]{IEEEtran}
\usepackage[hyphens]{url}
\usepackage[bookmarks=true]{hyperref}
\usepackage{fancyhdr}

\usepackage{graphicx} 
\usepackage{listings}
\usepackage{makecell}

\usepackage{multicol}
\usepackage[numbers]{natbib}
\usepackage{setspace}
\usepackage[table]{xcolor}
\usepackage{tabularx}
\usepackage{times}

\begin{document}

\pagestyle{fancy}
\fancypagestyle{firstpage}{%
  \chead{Presented at RSS 2024 SemRob Workshop}
}

\title{Language models are robotic planners: \\reframing plans as goal refinement graphs}


\author{\authorblockN{Ateeq Sharfuddin}
\authorblockA{School of Computer Science\\
Carnegie Mellon University\\
Pittsburgh, PA, USA\\
Email: asharfud@cs.cmu.edu}
\and
\authorblockN{Travis Breaux}
\authorblockA{School of Computer Science\\
Carnegie Mellon University\\
Pittsburgh, PA, USA\\
Email: breaux@cs.cmu.edu}}

\maketitle
\thispagestyle{firstpage}
\begin{abstract}
Successful application of large language models (LLMs) to robotic planning and execution may pave the way to automate numerous real-world tasks. Promising recent research has been conducted showing that the knowledge contained in LLMs can be utilized in making goal-driven decisions that are enactable in interactive, embodied environments. Nonetheless, there is a considerable drop in correctness of programs generated by LLMs. We apply goal modeling techniques from software engineering to large language models generating robotic plans. Specifically, the LLM is prompted to generate a step refinement graph for a task. The executability and correctness of the program converted from this refinement graph is then evaluated. The approach results in programs that are more correct as judged by humans in comparison to previous work.
\end{abstract}

\IEEEpeerreviewmaketitle

\section{Introduction}

Large language models (LLMs) are competitive against state-of-the-art models on challenging natural language processing (NLP) tasks, such as reading comprehension, question answering, and machine translation. Where the models have inherent limitations, such as solving arithmetic problems or formulating a response from recent events, researchers have demonstrated that LLMs can be augmented to use tools to compose responses \cite{schick2024toolformer}. In parallel, researchers have been studying whether the parametric\footnote{Some researchers also use the term semantic knowledge \cite{brohan2022can}.} knowledge contained in LLMs can be applied to the field of robotics. Specifically, can a robot act as an LLM’s ``hands and eyes" to perform a task while the LLM provides parametric knowledge about the task \cite{brohan2022can}? Successful application of LLMs to robotic planning and execution may pave the way to automate numerous real-world tasks.

Enabling robots to perform real-world tasks is often formulated as a planning problem \cite{xu2019regression}. Classic symbolic planners rely on predefined planning domains \cite{aeronautiques1998pddl, fikes1971strips}, which require manual effort to adapt to new environments \cite{xu2019regression}. A number of works have investigated applications of neural architectures to generalize robotic planning \cite{xu2018neural, xu2019regression}. Recently, research has shown that the knowledge encoded in LLMs about how to perform high-level tasks can be expanded into a series of low-level steps that are then actionable in interactive, embodied environments \cite{brohan2022can, huang2022language}: That said, there is a considerable drop in correctness in the programs generated by LLMs, as judged by humans \cite{huang2022language}.

In this paper, we report on findings from the application of goal modeling techniques from software engineering to robotic planning has any effect on the executability and correctness of programs generated by LLMs. Our contribution is both a method of adaptation and metrics for evaluation. We show how to reframe the problem of decomposing a task into steps to one of generating a goal refinement graph, which results in programs with increased correctness as judged by humans. We discuss the necessary preliminaries, including background on goal modeling techniques, the construction of the prompt, the dataset, and we present the overall approach in Section 3. The results of our evaluation are presented in Section 4. In Section 5, we discuss the results. Finally, we conclude and consider future research directions.

\section{Related Work}
In this section, we now review related work.

\textbf{Large language models} Previously, it had been shown that pretrained recurrent or transformer language models could be competitive against state-of-the-art models on numerous challenging NLP tasks such as reading comprehension, question answering, and machine translation without needing task-specific architectures\cite{brown2020language}. One major limitation with these approaches is that they still required task-specific datasets and task-specific fine-tuning. One observed trend was that increasing the capacity of the transformer models also increased the performance on downstream NLP tasks. Brown et al. tested this hypothesis by training a 175 billion parameter autoregressive language model called GPT-3, and they evaluate this model against two dozen NLP datasets \cite{brown2020language}. In their evaluations GPT-3 either set new state-of-the-art performance or outperformed other fine-tuned models \cite{brown2020language}. OpenAI introduced GPT-4 in 2023, which outperformed both previous LLMs and most state-of-the-art systems by a considerable margin on numerous benchmarks \cite{achiam2023gpt}. GPT-4 exhibits human-level performance in various professional and academic benchmarks \cite{achiam2023gpt}. Our experiments use GPT-3.5 Turbo, GPT-4, and GPT-4 Turbo models when generating sub-goals from high-level goals.

\textbf{Applications of LLMs to Robotics} Vemprala et al. investigated if and how the abilities of ChatGPT generalize to the domain of robotics \cite{vemprala2024chatgpt}. Unlike text-only applications, robotics requires understanding of real-world physics, the environmental context, and the ability to perform physical actions. These are beyond the scope of large language models, as the text needs to be translated into a logical sequence of physical actions. Vemprala et al. propose creation of a high-level function/wrapper library, which ChatGPT uses to write code to be deployed to the robot \cite{vemprala2024chatgpt}. Similarly, Yoshida et al. integrate GPT-4 with their humanoid-robot ``Alter3,” thereby allowing Alter3 to exhibit a pose when prompted in natural text \cite{yoshida2023text}. Yoshida et al. use two steps: 1) from a text prompt generate 10 lines of exaggerated descriptions, and 2) use the descriptions of joints’ motion direction (labeled as Axis in the paper) and these 10 lines of exaggerated output  to produce python code. This python code is transmitted to Alter3 to control the air compressors to observe Alter3 act out the pose \cite{yoshida2023text}. Liang et. al demonstrate that code-writing LLMs can be used to write new robot policy code given examples of natural language commands as comments followed by corresponding policy code \cite{liang2023code}.  Singh et al. follow a similar approach to Liang et. al and also use environment information for precondition checking given current state for plan executability \cite{singh2023progprompt}. These works are approaches using LLMs to generate code, which is then executed by the robots to enact scenarios.

\textbf{VirtualHome} Puig et al. offered a knowledge base of common household activities, or tasks, and all the steps needed by a robot to achieve each activity \cite{puig2018virtualhome}. Puig et al. implemented the most common actions in the Unity game engine, e.g., pick-up, switch-on/off, sit, etc. Unity’s physics, navigation, and kinematic models support the virtual agent to execute these programs in the simulated household environment \textit{VirtualHome}. Puig et al. used Unity’s NavMesh for navigation, which supports path planning that avoids obstacles, and RootMotion Final1K inverse kinematics package, which was used to animate the action to be performed by the agent \cite{puig2018virtualhome}. The evaluations in our paper uses a dataset derived from this knowledge base \cite{huang2022language}.

\textbf{LLMs for Planning} Huang et al. investigate if LLMs containing world-knowledge can successfully decompose a high-level command into low-level instructions suitable for robotic-execution \cite{huang2022language}. Huang et al. use GPT-3 and Puig et al.'s VirtualHome simulator. The plans produced were often not executable in the VirtualHome: The low-level instructions did not map precisely to admissible actions, left out common-sense actions, or contained linguistic ambiguities \cite{huang2022language}. Huang et al. considered two axes for evaluation: executability, whether an action plan can be correctly parsed and satisfies the common-sense constraints of the environment, and correctness, how similar a generated program is to human-written programs \cite{huang2022language}. Huang et al. used the longest common subsequence (LCS) between the generated program and the human-written program, normalized by the maximum length of the two programs. The maximum LCS is accepted when there are multiple human-written programs. Our work aims to achieve the same goal: Use LLMs to generate executable and correct programs, as judged by humans. In order to compare our results we use Huang et al.'s LCS evaluation metric, and also use Huang et al.'s dataset, which is a transformation of Puig et al.'s dataset \cite{huang2022language}.

\textbf{Requirements Engineering} The discipline to discover, understand, formulate, analyze, and agree on what problem needs to be solved, why it needs to be solved, and who are involved in solving the problem is requirements engineering (RE)\cite{lamsweerde2018requirements}, which is the first phase of modern software engineering. Requirements engineering can be used to investigate a machine's effect on the surrounding world and the assumptions we make about this world \cite{lamsweerde2018requirements}. We apply requirements engineering techniques to model how an embodied agent interacts with objects in the VirtualHome simulator.

\textbf{Language models of Code for graphs} Previous approaches modified the output format of the problem to solve graph problems with LLMs: The structure to be generated (e.g., a graph or a table) was ``serialized” into text, where serialization involved flattening the graph into a list of node pairs or into a specification language such as DOT \cite{madaan2022language}. LLMs struggle to generate these serializations given that LLMs are primarily trained on free-form text, and the serialized structured outputs strongly diverge from the majority of the training data \cite{madaan2022language}. Consequently, a language model trained on natural language text is likely to fail to capture the topology of the graph \cite{madaan2022language}. Madaan et al. argue and evaluate Code-LLMs, and demonstrate that Code-LLMs are better structured reasoners \cite{madaan2022language}. Madaan et al. perform evaluations against three types of tasks: script generation, entity state tracking, and explanation graph generation \cite{madaan2022language}. Our work is inspired by Madaan et al.'s approach: We use LLMs to represent our goal refinement graphs as code. To the best of our knowledge, our work is the first to use large language models to generate robotic plans with goal refinement graphs.

\section{Methods}
\subsection{Preliminaries}
We apply goal modeling to robotic planning. In robotic planning a program consists of a high-level command, called a \textit{task}, that is broken down into low-level instructions, called \textit{steps}, suitable for robotic execution \cite{brohan2022can, huang2022language,puig2018virtualhome}. An example program is shown in Listing \ref{list:program}. Puig et al. use programs to drive an artificial agent to perform tasks in a simulated household environment, the VirtualHome simulator \cite{puig2018virtualhome}. This VirtualHome simulator is built on top of the Unity3D game engine, which supplies the physics, navigation, and kinematic models \cite{puig2018virtualhome}. Huang et al. demonstrated that LLMs that embed world-knowledge can decompose a high-level command into an appropriate series of unambiguous low-level steps in this VirtualHome setting \cite{huang2022language}.

\begin{minipage}{0.95\linewidth}
\begin{lstlisting}[frame=single,language=C++,label=list:program, captionpos=b,caption=An example program with a task and steps.]
Task: Open bathroom window
Step 1: Walk to bathroom
Step 2: Walk to window
Step 3: Find window
Step 4: Open window
\end{lstlisting}
\end{minipage}

We investigate applying goal modeling techniques from requirements engineering to model robotic planning tasks. Specifically, we prompt an LLM to generate a refinement graph from a task. We show that the series of steps extracted from the refinement graph of a task is more accurate than prior work, as judged by humans \cite{huang2022language,puig2018virtualhome}.

\subsection{Goal model}
\begin{figure}[htb]
\begin{center}
\includegraphics[scale=0.3]{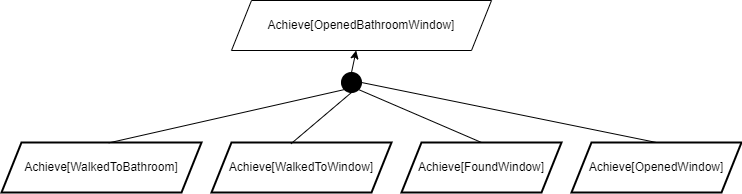}
\end{center}
\caption{An AND-refinement graph for the ``Open bathroom window" task.}
\label{fig:refinement}
\end{figure}

\begin{figure}[htb]
\begin{center}
\includegraphics[scale=0.3]{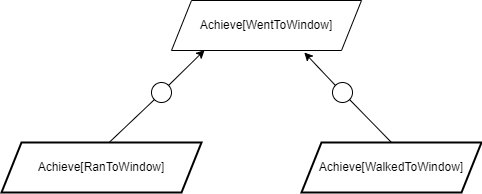}
\end{center}
\caption{An example OR-refinement graph.}
\label{fig:ORrefinement}
\end{figure}

A goal model in requirements engineering consists of a refinement graph illustrating how higher-level goals to be achieved, maintained or avoided by the system are refined into lower-level goals and, conversely, how lower-level goals contribute to higher-level goals \cite{lamsweerde2018requirements}. The lowest level goals in a refinement graph should be assignable to software-based agents that can satisfy those goals. A goal model is a directed, acyclic graph in which node represent goals that are connected by  directed edges, which represent asymmetric refinement relationships between goals  \cite{lamsweerde2018requirements}. Examples of AND- and OR-refinements are shown in Figure \ref{fig:refinement} and Figure \ref{fig:ORrefinement}, respectively. The goals are represented as parallelograms tilting right. The relationship of the contributing sub-goals to the parent goal is shown via directed edges traversing a refinement, which is represented with a small circle. A complete refinement, where no sub-goal is missing for the parent goal to be satisfied (i.e., all possible cases are covered), is represented as filled-in circle. An incomplete refinement is represented with an open circle and indicates that the refinement graph may not include all sub-goals necessary to satisfy the parent goal. When edges are joined into an AND-refinement it indicates how a high-level goal is decomposed into two or more lower-level sub-goals \cite{lamsweerde2018requirements}. In an AND-refinement, the parent goal is satisfied if and only if all sub-goals in the refinement are satisfied \cite{lamsweerde2018requirements}. In an OR-refinement, satisfying any alternative sub-goal results in the satisfaction of the parent goal. 

\begin{figure}[htb]
\begin{center}
\includegraphics[scale=0.3]{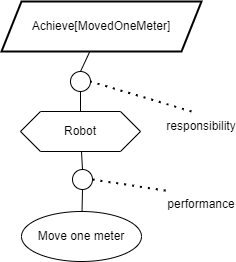}
\end{center}
\caption{Performance link with agent and operation and a goal.}
\label{fig:performance}
\end{figure}

An \textit{achievement goal} assigns expected behaviors where a target condition must sooner or later hold whenever some other condition holds in the current system state \cite{lamsweerde2018requirements}. An achievement goal refers to a future state from the current state, and the past participle of a suggestive verb is used for the goal’s name\cite{lamsweerde2018requirements}. An achievement goal for some \textit{TargetCondition} is written as \textit{Achieve[TargetCondition]}.

In goal modeling, an \textit{agent} can be considered the ``processor" that performs \textit{operations} to satisfy goals for which it is responsible \cite{lamsweerde2018requirements}. A system agent is represented with a hexagon and an operation is represented with an oval, as shown in Figure \ref{fig:performance}. In this regard, the virtual actor in the VirtualHome simulator is an agent and an action performed by this agent, i.e., the low-level step, is an operation. Each leaf goal has a performance link tying an agent to an operation, which when performed will satisfy the leaf achievement goal as shown in Figure \ref{fig:performance}. In this regard, one can observe that completing steps for a given task can be interpreted as achievement goals. 

For the robotic planning tasks we reframe the problem of decomposing a task into steps into a problem of generating a refinement graph of achievement sub-goals given a parent goal. Figure \ref{fig:refinement} represents the program in Listing \ref{list:program} as an AND-refinement of achievement goals, where each sub-goal is tied to a step and the parent goal is tied to the task. In other words, satisfying all the low-level achievement goals will result in the satisfaction of the high-level goal tied to the task. Consequently, generating these refinement graphs is framed as a code generation task, which was inspired by Madaan et al.'s work \cite{madaan2022language}.

\subsection{Prompt}
\label{subsec:prompt}
For code generation C++ was selected. C++ is strongly typed, a compiled language, and popular, leading to its usage in many open source projects. LLMs have been trained on a large number of open source repositories, and provide code completion for popular programming languages, including C++ \cite{githubnov2023, githubdocs2024,herring2023}. The LLM is prompted to complete a given code fragment. The code fragment is composed of a \textit{schema}, \textit{operations}, \textit{agents}, \textit{leaf goals}, a \textit{demonstration}, and a \textit{partial statement} for the LLM to complete, which we now discuss.

\subsubsection{Schema}
The schema consists of programming language idioms, such as class definitions and enumerations. These idioms represent edges, vertices, different types of goals, operations, actors, refinements, and different types of links. The schema contains all the elements necessary to represent refinement graphs in code. The schema is available in Appendix \ref{appendix:schema}.

\subsubsection{Operations}
The operations are actions that an agent may perform. An operation is directly correlated to a step. All the operations are listed in Appendix \ref{appendix:operations}.

\subsubsection{Agent}
There is only one agent, the virtual actor, who is a system agent. All the operations are performed by this one virtual actor. The agent is defined in Listing \ref{list:actor}. For the purposes of goal modeling, we consider this agent the machine affecting the surrounding world.

\begin{minipage}{0.95\linewidth}
\begin{lstlisting}[frame=single,language=C++,label=list:actor, captionpos=b,caption=An actor tied to the list of operations it can perform.]
Agent virtualPerson(
    "VirtualPerson",
    virtualPersonOperations);
\end{lstlisting}
\end{minipage}

\subsubsection{Leaf goals}
There is a leaf goal corresponding to each step. These leaf goals are all achievement goals since the target condition must hold sooner or later. Each leaf goal links an agent to the operation, which is the step that needs to be performed. In other words, the agent performing the operation will result in the satisfaction of the leaf goal. The leaf goals are specified in Appendix \ref{appendix:leafgoals}.

\subsubsection{Demonstration}
A single statement representing a refinement graph of a high-level goal refined into low-level goals is provided as a demonstration. This demonstration is in Listing \ref{list:demonstration}.

\begin{minipage}{0.95\linewidth}
\begin{lstlisting}[frame=single,language={C++},label=list:demonstration, captionpos=b,caption=An achievement goal demonstration for the LLM.]
AchieveGoal TurnedOffFloorLampInHomeOffice(
  "TurnedOffFloorLampInHomeOffice",
  {
    Refinement(
      AND_REFINEMENT,
      COMPLETE_REFINEMENT,
      {
        walkedToHomeOffice,
        walkedToFloorLampInHomeOffice,
        foundFloorLampInHomeOffice,
        switchedOffFloorLampInHomeOffice
      }
    )
  }
);
\end{lstlisting}
\end{minipage}

\subsubsection{Partial statement}
The last component of the code fragment is a partial statement wherein we start the definition of the achievement goal corresponding to the task as shown in Listing \ref{list:partial}. We follow the existing convention and use past participle tense to label the achievement goal \cite{lamsweerde2018requirements}. We revise the task to make it more precise and we write it as an achievement goal (explained under subsection \ref{subsec:approach}).

\begin{minipage}{0.95\linewidth}
\begin{lstlisting}[frame=single,language=C++,label=list:partial, captionpos=b,caption=Partial statement for an achievement goal]
  AchieveGoal TurnedOnFloorLampInHomeOffice(
\end{lstlisting}
\end{minipage}

The composed code fragment is provided as a prompt to GPT-4 (gpt-4-0613), GPT-4 Turbo (gpt-4-0125-preview), and GPT-3.5 Turbo (gpt-3.5-turbo-1106). The LLM completes the partial statement. The generated completion is a representation of the refinement graph as code as Listing \ref{list:response}. In this representation, the leaf goals to be satisfied correspond to the steps of the robot. The LLM response may use different language idioms than the ones used in the demonstration; however, the code returned represents a refinement graph.

\begin{minipage}{0.95\linewidth}
\begin{lstlisting}[frame=single,language=C++,label=list:response, captionpos=b, caption=LLM response depicting the goal refinement graph]
  "TurnedOnFloorLampInHomeOffice", 
  {
    Refinement(
      AND_REFINEMENT,
      COMPLETE_REFINEMENT,
      {
        walkedToHomeOffice,
        walkedToFloorLampInHomeOffice,
        foundFloorLampInHomeOffice,
        switchedOnFloorLampInHomeOffice
      }
    )
  }
);
\end{lstlisting}
\end{minipage}

The series of steps can now be generated from the performance links tied to the actor in each leaf node of this refinement graph. The implementation enumerating the refinement links and sub-goals, and retrieving the operation and producing the steps, is provided in the method \texttt{Goal::generateSteps} in Appendix \ref{appendix:schema}.

\subsection{Dataset}
We manually reviewed all 201 tasks in the dataset provided by Huang et al. \cite{huang2022language}. This dataset of programs is textual and was derived from a dataset originally written for Virtual Home, which is a visual 3D environment \cite{puig2018virtualhome}. We selected a subset of 20 tasks that are enactable by a robot. Additionally, our reasoning for avoiding many of the tasks follows similar reasoning provided by Sing et al. when they chose a narrower set of tasks from the dataset: some common sense actions for the objects are not available in VirtualHome\cite{singh2023progprompt} and missing from the programs. An example of this is the program for the task ``Open front door," which only has one step ``Walk to home office." Furthermore, some programs are duplicates under differing task names (for example, change TV channel and change TV channels). We discuss further in Section \ref{sec:discussion}.

For our dataset, the high-level achievement goal written for each task is revised to be more precise than the task in the task in Huang et al.'s dataset \cite{huang2022language}. For example, for the original task ``Turn on light" the achievement goal is written as \textit{Achieve[TurnedOnLightInBedRoom]}. Notably, there are multiple rooms with lights in the VirtualHome simulator but the dataset provided by Huang et al. is not exhaustive and does not contain programs to turn on lights in every room \cite{huang2022language}. Therefore, if the LLM were to generate a refinement graph where the actor turned on a light in a room for which no program exists in the dataset, the result will be incorrectly penalized by the evaluation metric.

\subsection{Approach}\label{subsec:approach}
For each task, a C++ code fragment as previously described in subsection \ref{subsec:prompt} is generated. The partial statement in this code fragment is updated to reflect the revised task written as an achievement goal. The code fragment is then transmitted as a user prompt to the LLM. The system prompt used is ``Output the next C++ line," with a sampling temperature of 0. The sampling temperature parameter for the OpenAI Chat completion API must be between 0 and 2, where lower values make the output more focused and deterministic \cite{openaichatcompletionapi2024}. The expectation is that the LLM responds with the completed code fragment, which contains the refinement graph. The series of steps can then be generated from the performance links tied to the actor in each leaf node in this refinement graph.

\subsection{Evaluation}
We evaluate the generated goal statements by computing the longest common subsequence (LCS) between the program generated by the LLM and the human-written program, normalized by the maximum length of the two programs. We select the maximum LCS when there are multiple crowd-sourced human-written programs for a single task, which was the evaluation metric proposed by Huang et al.\cite{huang2022language}.

\section{Results}
\begin{table}
\caption{The maximum normalized LCS of programs generated by GPT-4 Turbo, GPT-4, and GPT-3.5 Turbo.}
\label{table:results}
\begin{center}
\rowcolors{2}{white}{gray!20}
\newcolumntype{C}{>{\raggedright\let\newline\\\arraybackslash\hspace{0pt}}m{2cm}}
{\small %
\begin{tabularx}{\linewidth}{|p{1.35cm}|X|p{1cm}|p{0.9cm}|p{1.2cm}|}
\hline
\thead {Task} & 
\thead{\makecell[l]Achieve[Goal]} &
\thead{GPT4-T} & \thead{GPT4} & \thead{GPT3.5-T} \\
\hline
\makecell[l]{Turn on\\light} & \makecell[l]{TurnedOn\\FloorLamp\\InHomeOffice} & 1.0 & 1.0 & 1.0 \\
\makecell[l]{Turn on\\light} &
\makecell[l]{TurnedOnLight\\InDiningRoom} &
1.0 &
1.0 &
1.0 \\
\makecell[l]{Turn light\\off} &
\makecell[l]{TurnedOffLight\\InDiningRoom} &
1.0 &
1.0 &
1.0\\
\makecell[l]{Turn on\\light} &
\makecell[l]{TurnedOnLight\\InBedRoom} &
1.0 &
1.0 &
0.86 \\
\makecell[l]{Turn off\\light} &
\makecell[l]{TurnedOffLight\\InBedRoom} &
0.83 &
0.83 &
0.83 \\
\makecell[l]{Work} &
\makecell[l]{TurnedOn\\Computer\\InOfficeRoom} &
1.0 &
1.0 &
1.0 \\
\makecell[l]{Pick up\\phone} &
\makecell[l]{PickedUpPhone\\InDiningRoom} &
1.0 &
1.0 &
1.0 \\
Sleep &
\makecell[l]{WentToSleep\\InBedRoom} &
1.0 &
0.77 &
0.84 \\
\makecell[l]{Relax on\\sofa} &
SatOnCouch &
1.0 &
1.0 &
1.0 \\
Open window &
\makecell[l]{OpenedWindow\\InOfficeRoom} &
0.81 &
0.81 &
0.81 \\
\makecell[l]{Open\\bathroom\\window} &
\makecell[l]{OpenedWindow\\InBathRoom} &
1.0 &
1.0 &
1.0\\
Watch TV &
\makecell[l]{TurnedOn\\Television} &
1.0 &
1.0 &
1.0\\
\makecell[l]{Change TV\\channel} &
\makecell[l]{ChangedChannel\\WithRemote\\Control\\InHomeOffice} &
0.83 &
0.95 &
0.87 \\
\makecell[l]{Raise\\blinds} &
\makecell[l]{RaisedCurtains\\InOfficeRoom} &
0.66 &
0.61 &
0.71 \\
\makecell[l]{Sit in\\chair} &
\makecell[l]{SatInChair\\InDiningRoom} &
1.0 &
1.0 &
0.70\\
\makecell[l]{Go to\\toilet} &
SatOnToilet &
1.0 &
1.0 &
1.0\\
\makecell[l]{Take nap} &
TookNapOnBed &
1.0 &
1.0 &
0.63\\
\makecell[l]{Gaze out\\window} &
\makecell[l]{GazedOut\\Window\\InOfficeRoom} &
1.0 &
1.0 &
1.0\\
\makecell[l]{Sit} &
\makecell[l]{SatOnChair\\InDiningRoom} &
1.0 &
1.0 &
0.79\\
\makecell[l]{Pull up\\carpet} &
\makecell[l]{PulledMat\\InHomeOffice} &
0.6 &
0.85 &
0.85 \\
\hline

Percent &
&
93.66\% &
94.17\% &
89.44\% \\
\hline
\end{tabularx}
} 
\end{center}
\end{table}

Table \ref{table:results} shows the maximum normalized longest common subsequence (LCS) between the program generated by the LLMs and human-written program calculated using the evaluation metric described in subsection \ref{subsec:approach}. GPT4-Turbo generates programs that exactly match human-written programs in 15 of the 20 tasks (i.e., where LCS is 1.0). GPT-4 generates programs that exactly match human-written programs for 14 of the 20 tasks. Following Huang et al.’s methodology, we sum the LCS of all the tasks normalized by the sum of maximum possible LCS (20), to see overall LCS of 93.66\% and 94.17\% for the 20 programs generated by GPT-4 Turbo (gpt-4-0125-preview) and GPT-4 (gpt-4-0613), respectively. Similarly, GPT3.5-Turbo generates programs that exactly match human-written programs for 10 of the 20 tasks, and achieves an overall LCS of 89.44\%. The results are discussed in Section \ref{sec:discussion}.


\section{Discussion}
\label{sec:discussion}
The results indicate that GPT4 and GPT4-Turbo performed better than GPT-3.5 Turbo in generating programs as goal refinement graphs. For a majority of the selected tasks, GPT4 and GPT4-Turbo successfully refine each high-level goal into sub-goals, and the resulting programs exactly matches the human-written programs. For the cases where the resulting programs do not exactly match human-written programs, the LLMs identified most of the sub-goals for each high-level goal. All 20 generated programs have an LCS over 0.6. Prior work finds best overall LCS in the range between 31.05\% and  34.00\% \cite{huang2022language}. However, our overall LCS between 89.44\% and 93.66\% cannot be directly compared to prior work, since we selected a subset of 20 tasks from the dataset and instead of all the tasks. The prior work do not list individual LCS that we may have been able to compare against \cite{huang2022language, puig2018virtualhome}.

\subsection{Error analysis}\label{section:erroranalysis}

An exhaustive review of all programs where the normalized LCS was less than 1.0 follows. For the program to turn off lights in the bed room, where LCS was 0.83, all the LLMs added a sub-goal tied to the ``Find light" action. The ``Find [object]" step before performing another action with [object] is the common pattern for programs in the dataset. However, there is no ``Find light" step in the dataset for any program for turning off lights in the bedroom. Additionally, removing the sub-goal tied to finding the floor lamp step in the prompt's \textit{demonstration} results in LLMs not generating refinement graphs with these ``Find object" sub-goals. Consequently, the LLMs will generate refinement graphs where the resulting program now has an LCS of 1.0 for this task but removing this ``Find [object]" step from the \textit{demonstration} penalizes other programs in the experiment that do have a ``Find [object]" step. Similarly, for opening a window in the office room, the ``Find window" step is missing in all cases of human-written programs for the office room. However, all human-written programs for opening the window in the bathroom have a ``Find window" step. An option would be to utilize alternative OR-refinements to support both with and without ``Find object" steps, prompt the LLM multiple times with a larger sampling temperature such that we have refinement graphs both with and without these corresponding sub-goals, and an updated evaluation metric where we choose the maximum normalized LCS among all these returned responses.

For changing the TV channel, the program generated by GPT-4 achieved the highest normalized LCS of 0.95. We summarize how the generated programs differs from the program in the dataset. GPT-4 applied a sub-goal tied to ``Push remote control" instead of a sub-goal tied to ``Push button." GPT-4 Turbo essentially generated the same program as GPT-4 but missed the sub-goal tied to the ``Walk to home office" step. Finally, GPT-3.5 Turbo did not add a ``Find remote control" sub-goal before interacting with the remote control. 

For the program to raise blinds, GPT-4 misses sub-goals for both the ``Pull blind" steps. GPT-4 Turbo misses the sub-goal tied to ``Walk to home office" instead.

For pulling the mat, the human-written programs have a ``Touch mat" step but the corresponding sub-goal for this step is missing in the refinement graphs generated by GPT-4 Turbo, GPT-4, and GPT-3.5 Turbo. GPT-4.5 Turbo misses the sub-goal tied to the ``Walk to home office" step, as well.

For the program associated with sleeping, GPT-4 adds sub-goal tied to turning off the light whereas the human-written programs instead have a step to turn off the floor lamp instead of the light. The bed room has both a floor lamp and a light. Many other programs in the dataset involving the bedroom interact with the light instead of the floor lamp. GPT-3.5 Turbo does not include the sub-goals tied to the ``Find bed" and ``Sleep" steps. GPT-4 had the most rows with LCS of 1.0. Table \ref{table:missingsteps} summarizes steps added or missing from the programs generated by GPT-4 Turbo for tasks where LCS was below 1.0.

\begin{table}[htb!]
\caption{Steps missing or added in plans generated by GPT-4 Turbo}
\label{table:missingsteps}
\begin{center}
\rowcolors{2}{white}{gray!20}
\newcolumntype{C}{>{\raggedright\let\newline\\\arraybackslash\hspace{0pt}}m{3cm}}
\begin{tabular}{ |C|l|l|}
\hline
\thead{Achieve[Goal]} & \thead{Missing step} & \thead{Added step}\\
\hline
\makecell[l]{TurnedOffLight\\InBedRoom} & & Find light \\
\makecell[l]{OpenedWindow\\InOfficeRoom} & & Find window \\
\makecell[l]{ChangedChannel\\WithRemoteControl\\InHomeOffice} & Push button & \makecell[l]{Push remote\\control} \\
\makecell[l]{RaisedCurtains\\InOfficeRoom} & \makecell[l]{Walk to\\home office} &  \\
\makecell[l]{PulledMat\\InHomeOffice} & \makecell[l]{Walk to\\home office} & \\
\hline
\end{tabular}
\end{center}
\end{table}

\subsection{Threats to validity}
We manually reviewed all the programs in the dataset provided by Huang et al. Programs not relevant to robotic planning were excluded from the evaluation. Notably, if every sample for a program contained at least one step not relevant to accomplishing the task, the program was excluded. For example, all human-written samples for the task ``Turn on lights" (not the ``Turn on light" task) specific to the bedroom, contain two final steps: ``Find bed" and ``Lie on bed," none of which are necessary to turn on lights in the bedroom from a robotic planning perspective. Additionally, certain programs may not be executable in the VirtualHome simulator because the environment itself is an incomplete imitation of a physical home. If every human-written program for a task contains steps that cannot fulfill the task, the program was also excluded. For example, all samples for the task ``Open door" contain programs with only one step, ``Walk to home office." Obviously, this ``Walk to home office" step alone does not result in the virtual actor opening a door. Therefore, these programs were excluded. The remaining programs were modeled using achievement goals in an AND-refinement. We have threats to external validity: We do not claim that our results generalize to all the programs in the dataset. We also cannot claim that our results will generalize to other robotic planning datasets, either. We aim to apply goal modeling techniques to other robotic planning datasets in our future work.

\subsection{Additional concerns}
Because the VirtualHome environment itself supplies the physics, navigation, and kinematic models, the context and the present state of the virtual actor are not encoded in the high-level task \cite{huang2022language, puig2018virtualhome}. For example, for the task ``Turn on light," a possible sequence of steps annotated by a human were ``Walk to home office," ``Walk to floor lamp," and ``Switch on floor lamp." The environment provides the context for the virtual actor, such as, the current locations of both the actor and the light, and how to avoid obstacles and navigate the actor to the light via Unity's NavMesh \cite{puig2018virtualhome}. As a result, we did not need to consider modeling with maintenance or avoidance goals. All the programs could be modeled with just achievement goals in our work.

\section{Conclusion}
We presented applying goal modeling techniques from requirements engineering to the study of large language models generating robotic plans. We reframed the problem from decomposing a task into steps to a problem of generating a refinement graph. 20 programs were selected from the dataset provided by Huang et al. \cite{huang2022language}, and LLMs were used to generate refinement graphs for the tasks associated to these 20 programs. The resulting programs in the study were both more executable and correct in comparison to  prior work based on the LCS evaluation metric used by Huang et al. \cite{huang2022language}, achieving an overall maximum normalized LCS of 94.17\% with GPT-4. Given the success of this study, our next step is to to apply additional goal types and refinements to other robotic planning datasets in our future work.


\bibliographystyle{plainnat}
\bibliography{references}
\appendices
\section{Schema}
\label{appendix:schema}
\begin{lstlisting}[upquote=true, basicstyle=\footnotesize\ttfamily, language={C++}]
enum VertexType {
    NODE_TYPE_GOAL,
    NODE_TYPE_REFINEMENT,
    NODE_TYPE_OBSTACLE,
    NODE_TYPE_AGENT,
    NODE_TYPE_OPERATION
};

enum EdgeType
{
    REFINEMENT,
    RESPONSIBILITY,
    PERFORMANCE
};

enum OperationCategory
{
    ENVIRONMENT_OPERATION,
    SOFTWARE_TO_BE_OPERATION
};

enum GoalType
{
    BEHAVIORAL_GOAL,
    SOFT_GOAL
};

enum RefinementType
{
    AND_REFINEMENT,
    OR_REFINEMENT
};

enum BehavioralGoalType
{
    ACHIEVE_GOAL,
    MAINTAIN_GOAL,
};

enum SoftGoalType
{
    IMPROVE,
    INCREASE,
    MAXIMIZE,
    REDUCE,
    MINIMIZE
};

enum GoalCategory
{
    SATISFACTION,
    INFORMATION,
    STIMULUS_RESPONSE,
    ACCURACY,
    QOS_SAFETY,
    QOS_SECURITY_CONFIDENTIALITY,
    QOS_SECURITY_INTEGRITY,
    QOS_SECURITY_AVAILABILITY,
    QOS_PERFORMANCE_TIME,
    QOS_PERFORMANCE_SPACE
    // many more
};

struct Vertex
{
    explicit Vertex(VertexType type):type(type){}
    const VertexType type;
};

struct Edge
{
    explicit Edge(EdgeType type):type(type){}
    const EdgeType type;
};

struct Operation: Vertex
{
    explicit Operation(
            string name,
            OperationCategory category):
        Vertex(NODE_TYPE_OPERATION),
        name(name),
        category(category)
    {}

    virtual string toString() const {
        return "(" + name + ")";
    }

    const OperationCategory category;
    const string name;
};

struct Agent: Vertex
{
    explicit Agent(
            string name,
            list<Operation> performs):
        Vertex(NODE_TYPE_AGENT), 
        name(name),
        performs(performs)
    {}

    virtual string toString() const {
        return "<" + name + ">";
    }

    const string name;
    // operations the agent may perform
    const list<Operation> performs;
};

struct Goal;

struct Refinement: Vertex
{
    Refinement(RefinementType type,
               bool complete,
               list<Goal> subgoals):
        Vertex(NODE_TYPE_REFINEMENT),
        type(type),
        complete(complete),
        subgoals(subgoals)
    {
        if (subgoals.size() == 0) {
            throw ("refinement subgoals is empty");
        }
        if (type == OR_REFINEMENT &&
            subgoals.size() != 1) {
            throw ("OR refinement with one sub-goal");
        }
    }

    const bool complete; // complete refinement
    const RefinementType type;
    const list<Goal> subgoals;
};

struct PerformanceLink: Edge
{
    PerformanceLink(
            Agent & agent,
            Operation & operation):
        Edge(PERFORMANCE),
        agent(agent),
        operation(operation)
    {}

    const Agent & agent;
    const Operation & operation;
};

struct Goal: Vertex
{
    Goal(GoalType type,
         const string name,
         list<PerformanceLink> performs):
        Vertex(NODE_TYPE_GOAL),
        type(type),
        name(name),
        performs(performs)
    {}

    Goal(GoalType type,
         const string name,
         list<Refinement> refinements):
        Vertex(NODE_TYPE_GOAL),
        type(type),
        name(name),
        disjunctions(refinements)
    {}

    virtual string toString() const
    { return "\\" + name + "\\"; }

    virtual string toTree() const
    {
        string result = toString();
        result += "\n";
        for (auto & disjunction : disjunctions)
        {
            result += "|\n";
            result += "+";
            for (auto &subgoal : disjunction.subgoals)
            {
                result += subgoal.toTree();
            }
            result += "\n|\n+";
        }

        for (auto & perform : performs)
        {
            result += "- performs" + 
            perform.agent.toString() + "::" + 
            perform.operation.toString() + "\n";
        }

        return result;
    }

    virtual string generateSteps() const
    {
        string result;
        int i = 1;
        for (auto & refinement : disjunctions) {
            for (auto & subgoal : refinement.subgoals)
            {
                result += "Step " +
                                std::to_string(i++) +
                                ": ";
                result += subgoal
                                  .performs
                                  .begin()->operation.name;
                result += "\n";
            }
        }

        result += "\n";

        return result;
    }

    const string name;
    const GoalType type;
    const list<Refinement> disjunctions;
    const list<PerformanceLink> performs;
};

struct BehavioralGoal: Goal
{
    BehavioralGoal(const string name,
                   list<PerformanceLink> performs):
        Goal(BEHAVIORAL_GOAL, name, performs)
    {}

    BehavioralGoal(const string name,
                   list<Refinement> refinements):
        Goal(BEHAVIORAL_GOAL, name, refinements)
    {}
};

struct AchieveGoal: BehavioralGoal
{
    AchieveGoal(const string name,
                list<PerformanceLink> performs):
        BehavioralGoal(name, performs),
        type(ACHIEVE_GOAL) {}

    AchieveGoal(
            const string name,
            list<Refinement> refinements):
        BehavioralGoal(name, refinements),
        type(ACHIEVE_GOAL) {}

    virtual string toString() const
    { return "\\Achieve:" + name + "\\"; }

    const BehavioralGoalType type;
};

struct CeaseGoal: AchieveGoal // dual
{
    CeaseGoal(
            string name,
            list<PerformanceLink> performs):
        AchieveGoal(name, performs)
    {}

    CeaseGoal(
            string name,
            list<Refinement> refinements):
        AchieveGoal(name, refinements)
    {}

    virtual string toString() const
    { return "\\Cease:" + name + "\\";}
};

struct MaintainGoal: BehavioralGoal
{
    MaintainGoal(
            string name,
            list<PerformanceLink> performs):
        BehavioralGoal(name, performs),
        type(MAINTAIN_GOAL)
    {}

    MaintainGoal(
            string name,
            list<Refinement> refinements):
        BehavioralGoal(name, refinements),
        type(MAINTAIN_GOAL)
    {}

    virtual string toString() const
    { return "\\Maintain:" + name + "\\"; }

    const BehavioralGoalType type;
};


struct AvoidGoal: MaintainGoal // dual
{
    AvoidGoal(
            string name,
            list<PerformanceLink> performs):
        MaintainGoal(name, performs)
    {}

    AvoidGoal(
            string name,
            list<Refinement> refinements):
        MaintainGoal(name, refinements)
    {}

    virtual string toString() const
    { return "\\Avoid:" + name + "\\";}
};

struct SoftGoal: Goal
{
    SoftGoal(
            SoftGoalType type,
            string name,
            list<PerformanceLink> performs):
        Goal(SOFT_GOAL, name, performs),
        type(type)
    {}

    const SoftGoalType type;
};

struct ResponsibilityLink: Edge
{
    ResponsibilityLink():Edge(RESPONSIBILITY)
    {}
};
\end{lstlisting}

\section{Operations}
\label{appendix:operations}
\begin{lstlisting}[upquote=true, basicstyle=\footnotesize\ttfamily, language={C++}]
Operation findCup("Find cup",
    ENVIRONMENT_OPERATION);
Operation drinkCup("Drink cup",
    ENVIRONMENT_OPERATION);
Operation pourMilkIntoCup("Pour milk into cup",
    ENVIRONMENT_OPERATION);
Operation closeFreezer("Close freezer",
    ENVIRONMENT_OPERATION);
Operation openFreezer("Open freezer",
    ENVIRONMENT_OPERATION);
Operation grabMilk("Grab milk",
    ENVIRONMENT_OPERATION);
Operation findMilk("Find milk",
    ENVIRONMENT_OPERATION);
Operation walkToHomeOffice(
    "Walk to home office",
    ENVIRONMENT_OPERATION);
Operation walkToBedRoom("Walk to bedroom",
    ENVIRONMENT_OPERATION);
Operation walkToDiningRoom(
    "Walk to dining room",
    ENVIRONMENT_OPERATION);
Operation walkToBathRoom("Walk to bathroom",
    ENVIRONMENT_OPERATION);
Operation walkToBedInBedroom("Walk to bed",
    ENVIRONMENT_OPERATION);
Operation findBedInBedroom("Find bed",
    ENVIRONMENT_OPERATION);
Operation lieOnBed("Lie on bed",
    ENVIRONMENT_OPERATION);
Operation walkToFloorLampInHomeOffice(
    "Walk to floor lamp",
    ENVIRONMENT_OPERATION);
Operation findFloorLampInHomeOffice(
    "Find floor lamp",
    ENVIRONMENT_OPERATION);
Operation walkToLightInBedroom("Walk to light",
    ENVIRONMENT_OPERATION);
Operation findLightInBedroom("Find light",
    ENVIRONMENT_OPERATION);
Operation walkToLightInDiningRoom(
    "Walk to light",
    ENVIRONMENT_OPERATION);
Operation findLightInDiningRoom("Find light",
    ENVIRONMENT_OPERATION);
Operation walkToCouchInHomeOffice(
    "Walk to couch",
    ENVIRONMENT_OPERATION);
Operation findCouchInHomeOffice("Find couch",
    ENVIRONMENT_OPERATION);
Operation walkToDeskInHomeOffice("Walk to desk",
    ENVIRONMENT_OPERATION);
Operation walkToTelevisionInHomeOffice(
    "Walk to television",
    ENVIRONMENT_OPERATION);
Operation findTelevisionInHomeOffice(
    "Find television",
    ENVIRONMENT_OPERATION);
Operation walkToComputerInHomeOffice(
    "Walk to computer",
    ENVIRONMENT_OPERATION);
Operation findComputerInHomeOffice(
    "Find computer",
    ENVIRONMENT_OPERATION);
Operation switchOnFloorLampInHomeOffice(
    "Switch on floor lamp",
    ENVIRONMENT_OPERATION);
Operation switchOffFloorLampInHomeOffice(
    "Switch off floor lamp",
    ENVIRONMENT_OPERATION);
Operation switchOnLightInBedRoom(
    "Switch on light",
    ENVIRONMENT_OPERATION);
Operation switchOffLightInBedRoom(
    "Switch off light",
    ENVIRONMENT_OPERATION);
Operation switchOnLightInDiningRoom(
    "Switch on light",
    ENVIRONMENT_OPERATION);
Operation switchOffLightInDiningRoom(
    "Switch off light",
    ENVIRONMENT_OPERATION);
Operation sitOnCouch("Sit on couch",
    ENVIRONMENT_OPERATION);
Operation switchOnTelevisionInHomeOffice(
    "Switch on television",
    ENVIRONMENT_OPERATION);
Operation switchOffTelevisionInHomeOffice(
    "Switch off television",
    ENVIRONMENT_OPERATION);
Operation switchOnComputerInHomeOffice(
    "Switch on computer",
    ENVIRONMENT_OPERATION);
Operation switchOffComputerInHomeOffice(
    "Switch off computer",
    ENVIRONMENT_OPERATION);
Operation walkToPhoneInHomeOffice("Walk to phone",
    ENVIRONMENT_OPERATION);
Operation findPhoneInHomeOffice("Find phone",
    ENVIRONMENT_OPERATION);
Operation grabPhone("Grab phone",
    ENVIRONMENT_OPERATION);
Operation sleep("Sleep",
    ENVIRONMENT_OPERATION);
Operation walkToWindow1InHomeOffice(
    "Walk to window",
    ENVIRONMENT_OPERATION);
Operation findWindow1InHomeOffice(
    "Find window",
    ENVIRONMENT_OPERATION);
Operation openWindow1InHomeOffice("Open window",
    ENVIRONMENT_OPERATION);
Operation walkToWindow2InHomeOffice(
    "Walk to window",
    ENVIRONMENT_OPERATION);
Operation findWindow2InHomeOffice("Find window",
    ENVIRONMENT_OPERATION);
Operation openWindow2InHomeOffice("Open window",
    ENVIRONMENT_OPERATION);
Operation walkToWindowInBathroom("Walk to window",
    ENVIRONMENT_OPERATION);
Operation findWindowInBathroom("Find window",
    ENVIRONMENT_OPERATION);
Operation openWindowInBathroom("Open window",
    ENVIRONMENT_OPERATION);
Operation walkToWindowCurtain1InHomeOffice(
    "Walk to curtain",
    ENVIRONMENT_OPERATION);
Operation findWindowCurtain1InHomeOffice(
    "Find curtain",
    ENVIRONMENT_OPERATION);
Operation walkToWindowCurtain2InHomeOffice(
    "Walk to curtain",
    ENVIRONMENT_OPERATION);
Operation findWindowCurtain2InHomeOffice(
    "Find curtain",
    ENVIRONMENT_OPERATION);
Operation pullWindowCurtain1InHomeOffice(
    "Pull curtain",
    ENVIRONMENT_OPERATION);
Operation pullWindowCurtain2InHomeOffice(
    "Pull curtain",
    ENVIRONMENT_OPERATION);
Operation walkToChairInDiningRoom("Walk to chair",
    ENVIRONMENT_OPERATION);
Operation findChairInDiningRoom("Find chair",
    ENVIRONMENT_OPERATION);
Operation pullChairInDiningRoom("Pull chair",
    ENVIRONMENT_OPERATION);
Operation sitOnChairInDiningRoom("Sit on chair",
    ENVIRONMENT_OPERATION);
Operation walkToToiletInBathroom("Walk to toilet",
    ENVIRONMENT_OPERATION);
Operation findToiletInBathroom("Find toilet",
    ENVIRONMENT_OPERATION);
Operation sitOnToiletInBathroom("Sit on toilet",
    ENVIRONMENT_OPERATION);
Operation turnToWindow1InHomeOffice(
    "Turn to window",
    ENVIRONMENT_OPERATION);
Operation lookOutWindow1InHomeOffice(
    "Look at window",
    ENVIRONMENT_OPERATION);
Operation turnToWindow2InHomeOffice(
    "Turn to window",
    ENVIRONMENT_OPERATION);
Operation lookOutWindow2InHomeOffice(
    "Look at window",
    ENVIRONMENT_OPERATION);
Operation walkToRemoteControlInHomeOffice(
    "Walk to remote control",
    ENVIRONMENT_OPERATION);
Operation findRemoteControlInHomeOffice(
    "Find remote control",
    ENVIRONMENT_OPERATION);
Operation grabRemoteControlInHomeOffice(
    "Grab remote control",
    ENVIRONMENT_OPERATION);
Operation findButtonOnRemoteControlInHomeOffice(
    "Find button",
    ENVIRONMENT_OPERATION);
Operation pushButtonOnRemoteControlInHomeOffice(
    "Push remote control",
    ENVIRONMENT_OPERATION);
Operation putBackRemoteControlInHomeOffice(
    "Put back remote control",
    ENVIRONMENT_OPERATION);

list<Operation> virtualPersonOperations = {
        findCup, drinkCup, pourMilkIntoCup,
        closeFreezer, openFreezer, grabMilk,
        findMilk, walkToHomeOffice, walkToBedRoom,
        walkToDiningRoom, walkToBathRoom,
        walkToFloorLampInHomeOffice,
        walkToLightInBedroom,
        walkToLightInDiningRoom,
        walkToCouchInHomeOffice,
        walkToDeskInHomeOffice,
        walkToTelevisionInHomeOffice, 
        walkToComputerInHomeOffice,
        walkToBedInBedroom,
        switchOnFloorLampInHomeOffice,
        switchOffFloorLampInHomeOffice,
        switchOnLightInBedRoom,
        switchOffLightInBedRoom,
        switchOnLightInDiningRoom,
        switchOffLightInDiningRoom,
        sitOnCouch, switchOnTelevisionInHomeOffice,
        switchOffTelevisionInHomeOffice,
        switchOnComputerInHomeOffice,
        switchOffComputerInHomeOffice,
        walkToPhoneInHomeOffice, grabPhone, sleep,
        walkToWindow1InHomeOffice,
        openWindow1InHomeOffice,
        walkToWindowInBathroom, openWindowInBathroom,
        walkToWindowCurtain1InHomeOffice,
        walkToWindowCurtain2InHomeOffice,
        pullWindowCurtain1InHomeOffice,
        pullWindowCurtain2InHomeOffice,
        walkToChairInDiningRoom,
        pullChairInDiningRoom,
        sitOnChairInDiningRoom,
        walkToToiletInBathroom,
        sitOnToiletInBathroom,
        walkToRemoteControlInHomeOffice,
        findRemoteControlInHomeOffice,
        grabRemoteControlInHomeOffice,
        findButtonOnRemoteControlInHomeOffice,
        pushButtonOnRemoteControlInHomeOffice,
        putBackRemoteControlInHomeOffice,
        findFloorLampInHomeOffice,
        findLightInBedroom,
        findLightInDiningRoom,
        findCouchInHomeOffice,
        findTelevisionInHomeOffice,
        findComputerInHomeOffice,
        findPhoneInHomeOffice,
        findWindow1InHomeOffice,
        findWindow2InHomeOffice,
        findWindowInBathroom,
        findWindowCurtain1InHomeOffice,
        findWindowCurtain2InHomeOffice,
        findChairInDiningRoom,
        findToiletInBathroom,
        turnToWindow1InHomeOffice,
        lookOutWindow1InHomeOffice,
        turnToWindow2InHomeOffice,
        lookOutWindow2InHomeOffice, lieOnBed
    };

\end{lstlisting}
\section{Leaf goals}
\label{appendix:leafgoals}
\begin{lstlisting}[upquote=true, basicstyle=\footnotesize\ttfamily, language={C++}]
// virtual person performs the associated operation
// tied to a leaf goal leaf goals
AchieveGoal foundCup("FoundCup",
    { PerformanceLink(virtualPerson, findCup) });
AchieveGoal drankCup("DrankCup",
    { PerformanceLink(virtualPerson, drinkCup) });
AchieveGoal pouredMilkIntoCup("PouredMilkIntoCup",
    { PerformanceLink(
        virtualPerson,
        pourMilkIntoCup) });
AchieveGoal closedFreezer("ClosedFreezer",
    { PerformanceLink(
        virtualPerson,
        closeFreezer) });
AchieveGoal grabbedMilk("GrabbedMilk",
    { PerformanceLink(virtualPerson, grabMilk) });
AchieveGoal foundMilk("FoundMilk",
    { PerformanceLink(virtualPerson, findMilk) });
AchieveGoal grabbedPhone(
    "GrabbedPhone",
    { PerformanceLink(
        virtualPerson,
        grabPhone) });
AchieveGoal slept("Slept",
    { PerformanceLink(virtualPerson, sleep) });
AchieveGoal satOnCouch("SatOnCouch",
    { PerformanceLink(
        virtualPerson,
        sitOnCouch) });
AchieveGoal pulledChairInDiningRoom(
    "PulledChairInDiningRoom",
    { PerformanceLink(
        virtualPerson,
        pullChairInDiningRoom) });
AchieveGoal satOnChairInDiningRoom(
    "SatOnChairInDiningRoom",
    { PerformanceLink(
        virtualPerson,
        sitOnChairInDiningRoom) });
AchieveGoal walkedToToiletInBathroom(
    "WalkedToToiletInBathroom",
    { PerformanceLink(
        virtualPerson,
        walkToToiletInBathroom) });
AchieveGoal foundToiletInBathroom(
    "FoundToiletInBathroom",
    { PerformanceLink(
        virtualPerson,
        findToiletInBathroom) });
AchieveGoal satOnToiletInBathroom(
    "SatOnToiletInBathroom",
    { PerformanceLink(
        virtualPerson,
        sitOnToiletInBathroom) });
AchieveGoal grabbedRemoteControlInHomeOffice(
    "GrabbedRemoteControlInHomeOffice",
    { PerformanceLink(virtualPerson,
      grabRemoteControlInHomeOffice) });
AchieveGoal foundButtonOnRemoteControlInHomeOffice(
    "FoundButtonOnRemoteControlInHomeOffice",
    { PerformanceLink(virtualPerson,
      findButtonOnRemoteControlInHomeOffice) });
AchieveGoal 
  pushedButtonOnRemoteControlInHomeOffice(
    "PushedButtonOnRemoteControlInHomeOffice",
    { PerformanceLink(virtualPerson,
      pushButtonOnRemoteControlInHomeOffice) });
AchieveGoal placeBackRemoteControlInHomeOffice(
    "PlaceBackRemoteControlInHomeOffice",
    { PerformanceLink(virtualPerson,
      putBackRemoteControlInHomeOffice) });
AchieveGoal openedWindow1InHomeOffice(
    "OpenedWindow1InHomeOffice",
    { PerformanceLink(virtualPerson,
      openWindow1InHomeOffice) });
AchieveGoal openedWindow2InHomeOffice(
    "OpenedWindow2InHomeOffice",
    { PerformanceLink(virtualPerson,
      openWindow2InHomeOffice) });
AchieveGoal openedWindowInBathroom(
    "OpenedWindowInBathroom",
    { PerformanceLink(virtualPerson,
      openWindowInBathroom) });
AchieveGoal openedCurtain1InHomeOffice(
    "PulledCurtain1InHomeOffice",
    { PerformanceLink(virtualPerson,
      pullWindowCurtain1InHomeOffice) });
AchieveGoal openedCurtain2InHomeOffice(
    "PulledCurtain2InHomeOffice",
    { PerformanceLink(virtualPerson,
      pullWindowCurtain2InHomeOffice) });
AchieveGoal turnedToWindow1InHomeOffice(
    "TurnedToWindow1InHomeOffice",
    { PerformanceLink(virtualPerson,
      turnToWindow1InHomeOffice) });
AchieveGoal lookedOutWindow1InHomeOffice(
    "LookedOutWindow1InHomeOffice",
    { PerformanceLink(virtualPerson,
      lookOutWindow1InHomeOffice) });
AchieveGoal turnedToWindow2InHomeOffice(
    "TurnedToWindow2InHomeOffice",
    { PerformanceLink(virtualPerson,
      turnToWindow2InHomeOffice) });
AchieveGoal lookedOutWindow2InHomeOffice(
    "LookedOutWindow2InHomeOffice",
    { PerformanceLink(virtualPerson,
      lookOutWindow2InHomeOffice) });

// refinements
AchieveGoal foundAndDrankCup("FoundAndDrankCup",
    { Refinement(
        AND_REFINEMENT,
        true,
        {foundCup, drankCup}) });
AchieveGoal getSomethingToDrink(
    "GetSomethingToDrink",
    { Refinement(
        OR_REFINEMENT,
        true,
        {foundAndDrankCup}) });

// leaf goals: walked to a room
AchieveGoal walkedToHomeOffice(
    "WalkedToHomeOffice",
    { PerformanceLink(
        virtualPerson,
        walkToHomeOffice)
    });
AchieveGoal walkedToBedRoom(
    "WalkedToBedRoom",
    { PerformanceLink(
        virtualPerson,
        walkToBedRoom) });
AchieveGoal walkedToDiningRoom(
    "WalkedToDiningRoom",
    { PerformanceLink(
        virtualPerson,
        walkToDiningRoom) });
AchieveGoal walkedToBathRoom(
    "WalkedToBathRoom",
    { PerformanceLink(
        virtualPerson,
        walkToBathRoom) });

// leaf goals: walked to an object
// in a specific room
AchieveGoal walkedToBed(
    "WalkedToBedInBedRoom",
    { PerformanceLink(virtualPerson,
          walkToBedInBedroom) });
AchieveGoal foundBed(
    "FoundBedInBedRoom",
    { PerformanceLink(virtualPerson,
          findBedInBedroom) });
AchieveGoal walkedToCouchInHomeOffice(
    "WalkedToCouchInHomeOffice",
    { PerformanceLink(virtualPerson,
          walkToCouchInHomeOffice) });
AchieveGoal foundCouchInHomeOffice(
    "FoundCouchInHomeOffice",
    { PerformanceLink(virtualPerson,
          findCouchInHomeOffice) });
AchieveGoal walkedToTelevisionInHomeOffice(
    "WalkedToTelevisionInHomeOffice",
    { PerformanceLink(virtualPerson,
          walkToTelevisionInHomeOffice) });
AchieveGoal foundTelevisionInHomeOffice(
    "FoundTelevisionInHomeOffice",
    { PerformanceLink(virtualPerson,
          findTelevisionInHomeOffice) });
AchieveGoal walkedToDeskInHomeOffice(
    "WalkedToDeskInHomeOffice",
    { PerformanceLink(virtualPerson, 
          walkToDeskInHomeOffice) });
AchieveGoal walkedToFloorLampInHomeOffice(
    "WalkedToFloorLampInHomeOffice",
    { PerformanceLink(virtualPerson,
          walkToFloorLampInHomeOffice) });
AchieveGoal foundFloorLampInHomeOffice(
    "FoundFloorLampInHomeOffice",
    { PerformanceLink(virtualPerson,
          findFloorLampInHomeOffice) });
AchieveGoal walkedToComputerInHomeOffice(
    "WalkedToComputerInHomeOffice",
    { PerformanceLink(virtualPerson,
          walkToComputerInHomeOffice) });
AchieveGoal foundComputerInHomeOffice(
    "FoundComputerInHomeOffice",
    { PerformanceLink(virtualPerson,
          findComputerInHomeOffice) });
AchieveGoal walkedToLightInBedRoom(
    "WalkedToLightInBedRoom",
    { PerformanceLink(virtualPerson,
          walkToLightInBedroom) });
AchieveGoal foundLightInBedRoom(
    "FoundLightInBedRoom",
    { PerformanceLink(virtualPerson,
          findLightInBedroom) });
AchieveGoal walkedToLightInDiningRoom(
    "WalkedToLightInDiningRoom",
    { PerformanceLink(virtualPerson,
          walkToLightInDiningRoom) });
AchieveGoal foundLightInDiningRoom(
    "FoundLightInDiningRoom",
    { PerformanceLink(virtualPerson,
          findLightInDiningRoom) });
AchieveGoal walkedToChairInDiningRoom(
    "WalkedToChairInDiningRoom",
    { PerformanceLink(virtualPerson,
          walkToChairInDiningRoom) });
AchieveGoal foundChairInDiningRoom(
    "FoundChairInDiningRoom",
    { PerformanceLink(virtualPerson,
          findChairInDiningRoom) });
AchieveGoal walkedToPhoneInHomeOffice(
    "WalkedToPhoneInHomeOffice",
    { PerformanceLink(virtualPerson,
          walkToPhoneInHomeOffice) });
AchieveGoal foundPhoneInHomeOffice(
    "FoundPhoneInHomeOffice",
    { PerformanceLink(virtualPerson,
          findPhoneInHomeOffice) });
AchieveGoal walkedToWindow1InHomeOffice(
    "WalkedToWindow1InHomeOffice",
    { PerformanceLink(virtualPerson,
          walkToWindow1InHomeOffice) });
AchieveGoal foundWindow1InHomeOffice(
    "FoundWindow1InHomeOffice",
    { PerformanceLink(virtualPerson,
          findWindow1InHomeOffice) });
AchieveGoal walkedToWindow2InHomeOffice(
    "WalkedToWindow2InHomeOffice",
    { PerformanceLink(virtualPerson,
          walkToWindow2InHomeOffice) });
AchieveGoal foundWindow2InHomeOffice(
    "FoundWindow2InHomeOffice", 
    { PerformanceLink(virtualPerson,
          findWindow2InHomeOffice) });
AchieveGoal walkedToWindowInBathroom(
    "WalkedToWindowInBathroom",
    { PerformanceLink(virtualPerson,
          walkToWindowInBathroom) });
AchieveGoal foundWindowInBathroom(
    "FoundWindowInBathroom",
    { PerformanceLink(virtualPerson,
          findWindowInBathroom) });
AchieveGoal walkedToCurtain1InHomeOffice(
    "WalkedToCurtain1InHomeOffice",
    { PerformanceLink(virtualPerson,
          walkToWindowCurtain1InHomeOffice) });
AchieveGoal foundCurtain1InHomeOffice(
    "FoundCurtain1InHomeOffice",
    { PerformanceLink(virtualPerson,
          findWindowCurtain1InHomeOffice) });
AchieveGoal walkedToCurtain2InHomeOffice(
    "WalkedToCurtain2InHomeOffice",
    { PerformanceLink(virtualPerson,
          walkToWindowCurtain2InHomeOffice) });
AchieveGoal foundCurtain2InHomeOffice(
    "FoundCurtain2InHomeOffice",
    { PerformanceLink(virtualPerson,
          findWindowCurtain2InHomeOffice) });
AchieveGoal walkedToRemoteControlInHomeOffice(
    "WalkedToRemoteControlInHomeOffice",
    { PerformanceLink(virtualPerson,
          walkToRemoteControlInHomeOffice) });
AchieveGoal foundRemoteControlInHomeOffice(
    "FoundRemoteControlInHomeOffice",
    { PerformanceLink(virtualPerson,
          findRemoteControlInHomeOffice) });

// leaf goals: performed an action on an object
// in a specific room
AchieveGoal liedOnBedInBedRoom(
    "LiedOnBedInBedRoom",
    { PerformanceLink(
         virtualPerson,
         lieOnBed) });
AchieveGoal switchedOnTelevisionInHomeOffice(
    "SwitchedOnTelevisionInHomeOffice",
    { PerformanceLink(
        virtualPerson,
        switchOnTelevisionInHomeOffice) });
AchieveGoal switchedOffTelevisionInHomeOffice(
    "SwitchedOffTelevisionInHomeOffice",
    { PerformanceLink(virtualPerson,
          switchOffTelevisionInHomeOffice) });
AchieveGoal grabbedPhoneInHomeOffice(
    "GrabbedPhoneInHomeOffice",
    { PerformanceLink(virtualPerson, grabPhone) });
AchieveGoal switchedOnFloorLampInHomeOffice(
    "SwitchedOnFloorLampInHomeOffice",
    { PerformanceLink(virtualPerson,
          switchOnFloorLampInHomeOffice) });
AchieveGoal switchedOffFloorLampInHomeOffice(
    "SwitchedOffFloorLampInHomeOffice",
    { PerformanceLink(virtualPerson,
          switchOffFloorLampInHomeOffice) });
AchieveGoal switchedOnLightInBedRoom(
    "SwitchedOnLightInBedRoom",
    { PerformanceLink(virtualPerson,
          switchOnLightInBedRoom) });
AchieveGoal switchedOffLightInBedRoom(
    "SwitchedOffLightInBedRoom",
    { PerformanceLink(virtualPerson,
          switchOffLightInBedRoom) });
AchieveGoal switchedOnLightInDiningRoom(
    "SwitchedOnLightInDiningRoom",
    { PerformanceLink(virtualPerson,
          switchOnLightInDiningRoom) });
AchieveGoal switchedOffLightInDiningRoom(
    "SwitchedOffLightInDiningRoom",
    { PerformanceLink(virtualPerson,
          switchOffLightInDiningRoom) });
AchieveGoal switchedOnComputerInHomeOffice(
    "SwitchedOnComputerInHomeOffice",
    { PerformanceLink(virtualPerson,
          switchOnComputerInHomeOffice) });
AchieveGoal switchedOffComputerInHomeOffice(
    "SwitchedOffComputerInHomeOffice",
    { PerformanceLink(virtualPerson,
          switchOffComputerInHomeOffice) });

\end{lstlisting}
\twocolumn
\end{document}